\documentclass{article}

\usepackage{arxiv}

\usepackage[utf8]{inputenc} % allow utf-8 input
\usepackage[T1]{fontenc}    % use 8-bit T1 fonts
\usepackage{hyperref}       % hyperlinks
\usepackage{url}            % simple URL typesetting
\usepackage{booktabs}       % professional-quality tables
\usepackage{amsfonts}       % blackboard math symbols
\usepackage{nicefrac}       % compact symbols for 1/2, etc.
\usepackage{microtype}      % microtypography
\usepackage{graphicx}
\usepackage{natbib}
\usepackage{doi}
\usepackage{amsmath}
\usepackage{algorithm}
\usepackage{algpseudocode}
\usepackage{caption}
\usepackage{subcaption}
\usepackage{comment}

\usepackage[table, xcdraw]{xcolor}
\usepackage{todonotes}
\usepackage{caption}
\usepackage{subcaption}
\usepackage{multirow}

\title{A condensing approach to multiple shooting neural ordinary differential equation}

\author{ 
    {\hspace{1mm}Siddharth Prabhu}\\
	Department of Chemical \\ and Biomolecular Engineering\\
	Lehigh University\\
	Bethlehem, PA 18015 \\
	\texttt{scp220@lehigh.edu} \\
	\And
    {\hspace{1mm}Srinivas Rangarajan} \\
	Department of Chemical \\ and Biomolecular Engineering\\
	Lehigh University\\
	Bethlehem, PA 18015 \\
	\texttt{srr516@lehigh.edu} \\
        \And
    {\hspace{1mm}Mayuresh Kothare} \\
	Department of Chemical \\ and Biomolecular Engineering\\
	Lehigh University\\
	Bethlehem, PA 18015 \\
	\texttt{mvk2@lehigh.edu} \\
}

\renewcommand{\shorttitle}{}

\begin{document}
\maketitle

\begin{abstract}
    Multiple-shooting is a parameter estimation approach for ordinary differential equations. In this approach, the trajectory is broken into small intervals, each of which can be integrated independently. Equality constraints are then applied to eliminate the shooting gap between the end of the previous trajectory and the start of the next trajectory. Unlike single-shooting, multiple-shooting is more stable, especially for highly oscillatory and long trajectories. In the context of neural ordinary differential equations, multiple-shooting is not widely used due to the challenge of incorporating general equality constraints. In this work, we propose a condensing-based approach to incorporate these shooting equality constraints while training a multiple-shooting neural ordinary differential equation (MS-NODE) using first-order optimization methods such as Adam.
\end{abstract}

% keywords can be removed
\keywords{Multiple-Shooting \and NeuralODE \and Equality Constraint Neural Network \and Automatic Differentiation}

\section{Introduction}

\begin{figure}[h!]
    \begin{subfigure}{0.48\linewidth}
        \centering
        % [trim={left bottom right top},clip]
        \includegraphics[width = \linewidth, height = 0.26\textheight, trim = 150 120 150 120, clip]{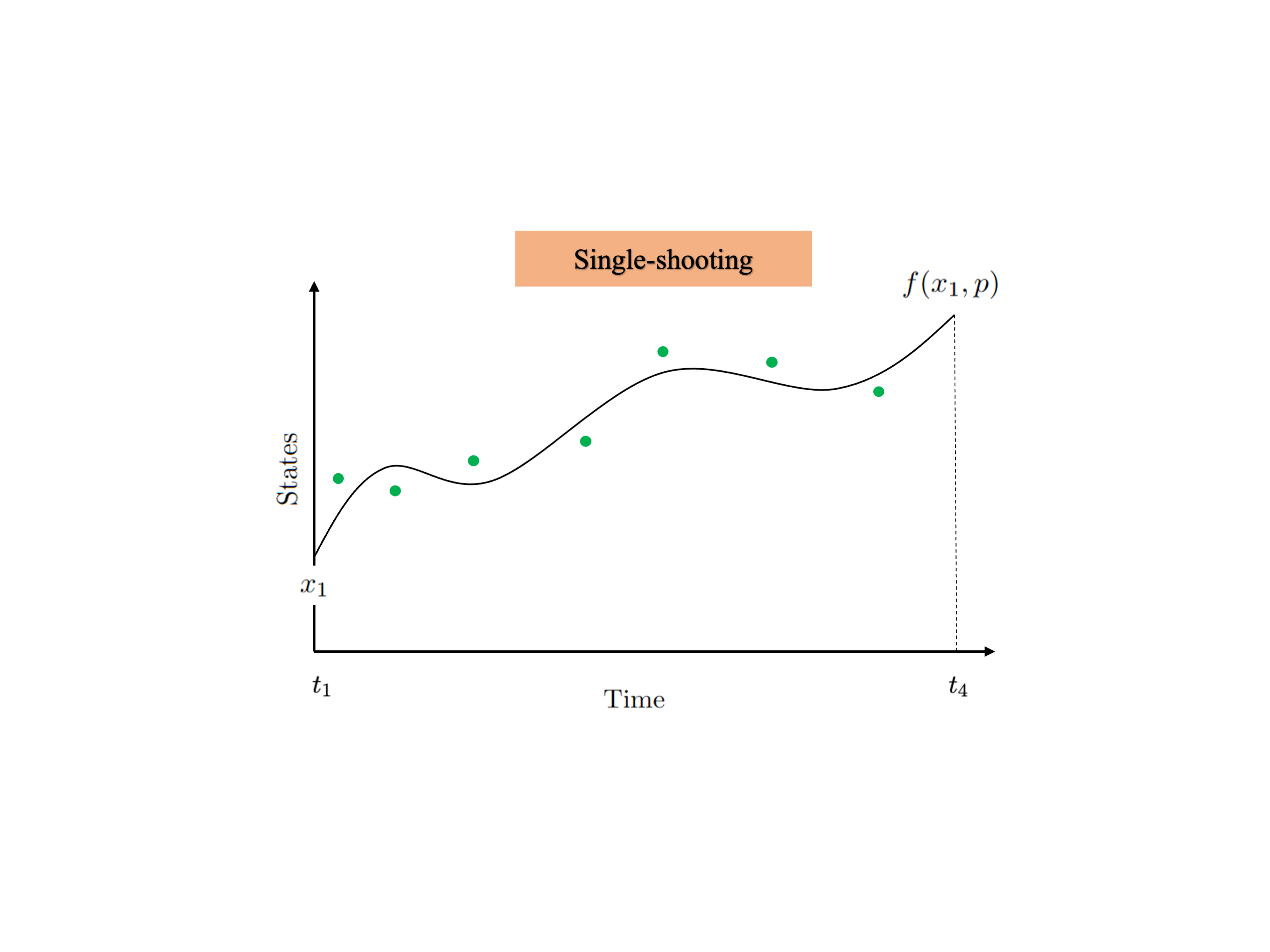}
        \caption{}
        \label{fig:single-shooting}
    \end{subfigure}
    \centering
    \begin{subfigure}{0.48\linewidth}
        \centering
        % [trim={left bottom right top},clip]
        \includegraphics[width = \linewidth, height = 0.26\textheight, trim = 130 120 130 120, clip]{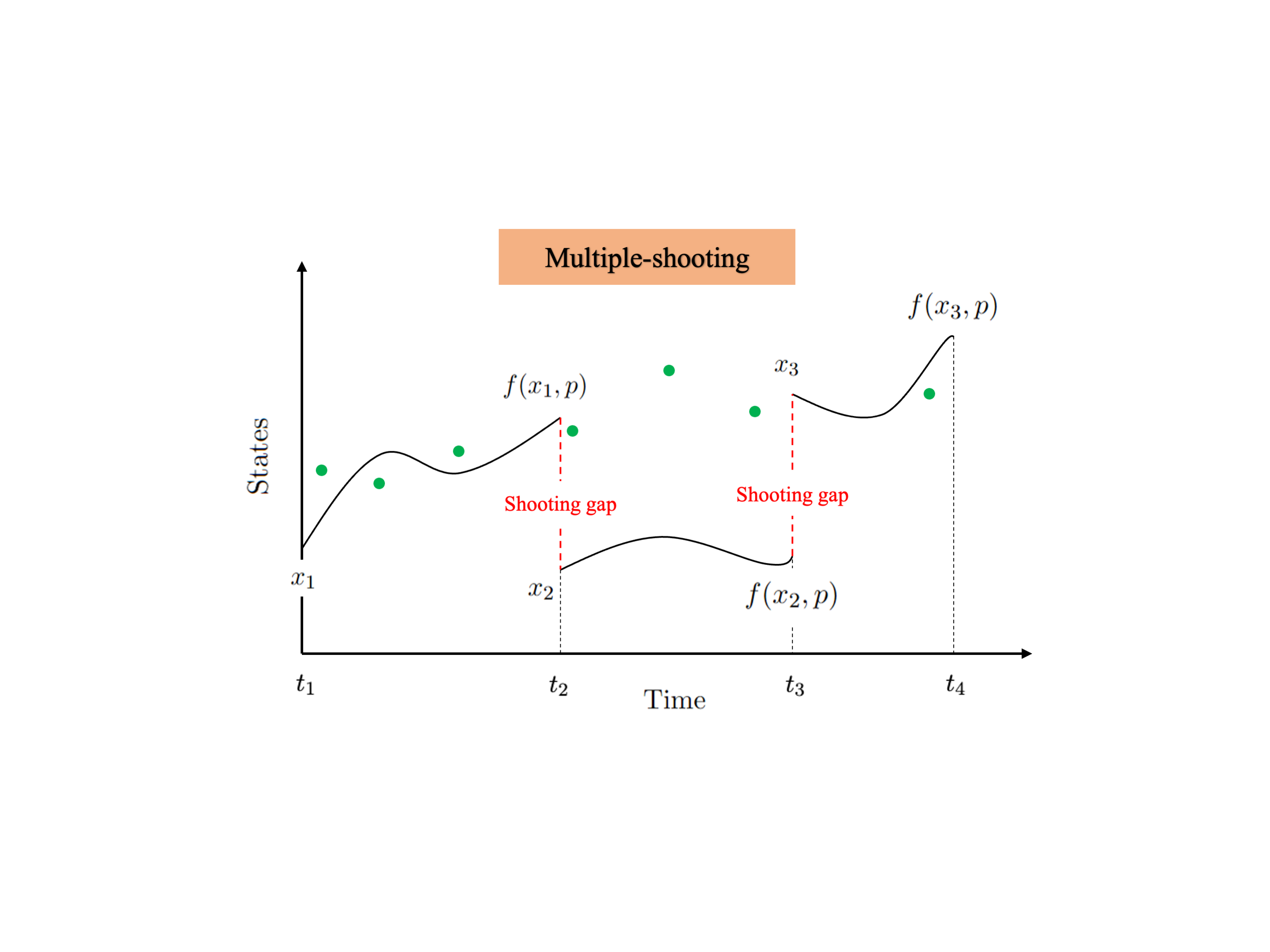}
        \caption{}
        \label{fig:multiple-shooting}
    \end{subfigure}
    \caption{A pictorial representation of single-shooting (a) and multiple-shooting (b) in the context of parameter estimation of ordinary differential equations. The green points are measurements, while the solid black line shows the models prediction.}
    \label{fig:shooting}
\end{figure}

Neural ordinary differential equation (NODE) \citep{chen2018neural, rackauckas2020universal} approximates the dynamics of the system as a neural network, which is integrated to get the trajectory of the states. The parameters of NODE are obtained by solving an optimization problem using a single-shooting approach \citep{vassiliadis1994solution1, vassiliadis1994solution2}. Single-shooting approach, however, is difficult to converge and unstable, especially for long and highly complex oscillatory dynamics. Multiple-shooting addresses these issues by dividing the trajectory into small intervals, which can be handled in parallel, while adding continuity equations as equality constraints to the optimization problem \citep{bock1984multiple, Diehl2006-cm}. A pictorial representation of the different shooting approaches is shown in Figure \ref{fig:shooting}. While these additional equality constraints can easily be incorporated in a typical Newton-based nonlinear optimization solver, incorporating them while training a neural network using first-order optimization techniques is not straightforward. 

Previous work by \cite{massaroli2021differentiable} incorporates multiple-shooting as differentiable layers in a neural network. However, their main focus is on solving forward simulations in parallel using a modification of the parareal method \citep{maday2002parareal}, while the training of the neural network parameters is still performed using a single-shooting approach. \cite{9651533} solves a constraint multiple shooting problem by reformulating it as an unconstrained augmented Lagrangian problem and applying either first-order or quasi-second-order optimization techniques. \cite{bradley2021two} introduces a multiple-shooting-like approach in which each shooting interval begins with a measurement used as the initial condition, thereby removing the need for explicit equality constraints. While they were able to successfully train such a NODE model on the training data, they were unable to extrapolate the model to unseen test data. Additionally, when the measurements are noisy, such an approach may break down because the endpoint of the previous shooting interval may not align with the noisy measurement, which serves as the initial condition of the next shooting interval. Some other works \citep{chen2024physics, rangarajan2022expressing, PhysRevLett.126.098302}, though not in the context of multiple shooting, have incorporated equality constraints during neural network training, a formulation to which the multiple-shooting problem can be reduced. However, these approaches are limited to either linear or linearly separable equality constraints, both of which are unlikely to arise in typical multiple-shooting scenarios. In this paper, we take a condensing-based \citep{bock1984multiple, albersmeyer2010lifted} approach for training NODE subject to continuity constraints (shooting gap equals zero) arising from a multiple-shooting formulation. This procedure ensures that the updates to the continuity variables ($\Delta x$) and the parameters ($\Delta p$) satisfy the first-order Taylor expansion of the equality constraints at each iteration. We demonstrate the applicability of this algorithm on several complex oscillatory dynamical systems and compare its performance with that of a model trained using naive single-shooting.

\section{Method}

In MS-NODE, we want to solve the following optimization problem 

\begin{align}
\label{}
\begin{split}
    & \min _{x_1, x_2, \cdots x_m, p} \ \sum _{k = 1}^{m} \phi (x_k, p) \\
    \text{subject to} & \\
    & \quad x_1 = \hat{x}_1 \\
    & \quad x_{k + 1} = x_{k} + \int _{t_K}^{t_{k + 1}} f(x_k, p) \quad \text{for} \quad k = 1, 2, \cdots, m - 1
\end{split}
\end{align}

\noindent where $x_k \in \mathbb{R}^n$ are the states, $\hat{x}_0$ is the measured initial condition, $\phi$ is the loss function and the equality constraints are the boundary constraints for each multiple shooting block. Let 

\begin{equation}
    F_k(x_k, p) = x_k + \int _{t_k}^{t_{k + 1}} f(x_k, p), \quad G = 
    \begin{bmatrix}
        x_1 - \hat{x}_1 \\
        x_2 - F_1(x_1, p) \\
        \vdots \\ 
        x_m - F_{m - 1}(x_{m - 1}, p)
    \end{bmatrix} _{mn \times 1}, \quad \text{and} \quad \Phi = \sum _{i = 1}^m \phi (x_i, p)
\end{equation}

\noindent where $G$ is a matrix with $m$ constraints and each constraint has dimension $n$, $F_k$ can be obtained using appropriate time-stepping integration schemes. For convenience, we will drop the brackets in front of $F_k$. The Lagrangian of the optimization problem in equation \ref{eqn:lagrangian} is written as 

\begin{equation} \label{eqn:lagrangian}
    L(x, p, \lambda ) = \Phi + \lambda ^T G
\end{equation}

\noindent and the KKT conditions are as follows 

\begin{align}\label{eqn:kkt}
\begin{split}
    & L_x = \Phi _x + \lambda ^T G_x = 0 \\
    & L_p = \Phi _p + \lambda ^T G_p = 0 \\
    & L_{\lambda} = G = 0
\end{split}
\end{align}

where $L_x \in \mathbb{R}^{mn \times 1}, L_p \in \mathbb{R}^{p \times 1}, L_{\lambda} \in \mathbb{R}^{mn \times 1}$ is the partial derivative of $L$ with respect to $x$, $p$, and $\lambda$ respectively. Similarly $G_x \in \mathbb{R}^{mn \times mn}, G_p \in \mathbb{R}^{mn \times p} $ is the partial derivative of $G$ with respect to $x$, and $p$ respectively. The Newton step for the system of equations in \ref{eqn:kkt} is as follows 

\begin{equation} \label{eqn:newton}
    \begin{bmatrix}
        L_{xx} & L_{xp} & G_x^T \\
        L_{px} & L_{pp} & G_p^T \\
        G_x & G_p & 0 \\ 
    \end{bmatrix} 
    \begin{bmatrix}
        \Delta x \\
        \Delta p \\
        \Delta \lambda \\
    \end{bmatrix} = - 
    \begin{bmatrix}
        L_x \\
        L_p \\
        G \\
    \end{bmatrix}
\end{equation}

Note that while training a neural network second-order optimization methods such as the Newton method are not used. In case of first-order optimization methods such as Gradient descent, which is often used for training a neural network, equation \ref{eqn:newton} can be rewritten as 

\begin{equation} 
    \begin{bmatrix}
        I_{xx} & 0 & G_x^T \\
        0 & I_{pp} & G_p^T \\
        G_x & G_p & 0 \\ 
    \end{bmatrix} 
    \begin{bmatrix}
        \Delta x \\
        \Delta p \\
        \Delta \lambda \\
    \end{bmatrix} = - 
    \begin{bmatrix}
        L_x \\
        L_p \\
        G \\
    \end{bmatrix}
\end{equation}

where $I_{xx} \in \mathbb{R}^{mn \times mn}$, $I_{pp} \in \mathbb{R}^{p \times p}$ are identity matrix and $0$ is a zero matrix of appropriate dimensions. Solving the system of equations gives

\begin{align}\label{eqn:x_update}
\begin{split}
    & I_{xx} \Delta x + G_x^T \Delta \lambda = - L_x \\
    & \Delta x = - \left[ G_x^T \Delta \lambda + L_x \right] \\
\end{split}
\end{align}

\begin{align}\label{eqn:p_update}
\begin{split}
    & I_{pp} \Delta p + G_p^T \Delta \lambda = - L_p \\
    & \Delta p = - \left[ G_p^T \Delta \lambda + L_p \right]
\end{split}
\end{align}

\begin{align}\label{eqn:lambda_update}
\begin{split}
    & G_x \Delta x + G_p \Delta p = - G \\
    & - G_x \left[ G_x^T \Delta \lambda + L_x \right] - G_p \left[ G_p^T \Delta \lambda + L_p \right] = - G \\
    & \Delta \lambda = \left[ G_xG_x^T + G_p G_p^T \right]^{-1} \left[ G - G_x L_x - G_p L_p \right]
\end{split}
\end{align}

Substituting for $\Delta \lambda$ in equation \ref{eqn:x_update} and \ref{eqn:p_update} gives 

\begin{align}\label{eqn:final_update}
\begin{split}
    & \Delta x = - \left[ G_x^T \left[ G_xG_x^T + G_p G_p^T \right]^{-1} \left[ G - G_x L_x - G_p L_p \right] + L_x \right]  \\
    & \Delta p = - \left[ G_p^T \left[ G_xG_x^T + G_p G_p^T \right]^{-1} \left[ G - G_x L_x - G_p L_p \right] + L_p \right] \\
\end{split}
\end{align}

Instead of calculating the matrix explicitly and then taking its inverse in equation \ref{eqn:lambda_update}, we use the conjugate gradient method, which only uses the cheaper Hessian-vector products and can be calculated efficiently. To do so, we introduce four new algorithms to efficiently deal with Jacobian-vector-product and vector-Jacobian-product of $G_x$, and $G_p$. Note that each of these algorithms can exploit GPU's by using primitives defined in \cite{wen2022programming}

\begin{algorithm}
\caption{Computing $ \left[ G_p v \right]$ }
\begin{algorithmic}
    \Require Tangent vector $ v \in \mathbb{R}^{p \times 1}$ 
    \State $Y \gets \text{Empty}(v) \in \mathbb{R}^{m \times n}$ \Comment{Initialize empty matrix $Y$}
    \State $ Y[1] \gets 0 $ \Comment{Initialize the first row of matrix to zero}
    \For{$k = 2, 3, \cdots, m$}
    \State $Y[k] \gets \frac{\partial F_{k - 1}}{\partial p} v $
    \EndFor
    \State $y \gets \text{Reshape}(Y) \in \mathbb{R}^{mn \times 1}$ \Comment{Vectorize the matrix}
    \State \Return $y$
\end{algorithmic}
\label{algo:gpv}
\end{algorithm}

\begin{algorithm}
\caption{Computing $ \left[ v^T G_p \right]$ }
\begin{algorithmic}
    \Require Co-tangent vector $ v \in \mathbb{R}^{mn \times 1}$ 
    \State $ V \gets \text{Reshape}(v) \in \mathbb{R}^{m \times n}$
    \State $ y \gets 0 \in \mathbb{R}^{p \times 1}$ \Comment{Initialize empty vector $y$}
    \For{$k = 2, 3, \cdots, m$}
    \State $y \gets y +  V[k]^T \frac{\partial F_{k-1}}{\partial p} $
    \EndFor
    \State \Return $y$
\end{algorithmic}
\label{algo:vgp}
\end{algorithm}

\begin{algorithm}
\caption{Computing $ \left[ G_x v \right]$}
\begin{algorithmic}
    \Require Tangent vector $v \in \mathbb{R}^{mn \times 1}$
    \State $ V \gets \text{Reshape}(v) \in \mathbb{R}^{m \times n}$
    \State $ Y[1] \gets V[1]$ \Comment{Initialize the first row of the matrix}
    \For{$k = 2, 3, \cdots, m$} \Comment{Forward pass}
    \State $Y[k] \gets V[k] - \frac{\partial F_{k-1}}{\partial x_{k-1}} Y[k - 1] $
    \EndFor
    \State $y \gets \text{Reshape}(Y) \in \mathbb{R}^{mn \times 1}$ \Comment{Vectorize the matrix}
    \State \Return $y$
\end{algorithmic}
\label{algo:gxv}
\end{algorithm}

\begin{algorithm}
\caption{Computing $ \left[ v^T G_x \right] $}
\begin{algorithmic}
    \Require Co-tangent vector $v \in \mathbb{R}^{mn \times 1}$
    \State $ V \gets \text{Reshape}(v) \in \mathbb{R}^{m \times n}$
    \State $Y \gets \text{Empty}(v) \in \mathbb{R}^{m \times n}$ \Comment{Initialize empty matrix $Y$}
    \State $ Y[m] \gets - V[m]$ \Comment{Initialize the last row of the matrix}
    \For{$k = m, m - 1, \cdots, 2$} \Comment{Backward pass}
    \State $Y[k - 1] \gets V[k - 1] - Y[k]^T \frac{\partial F_{k-1}}{\partial x_{k - 1}} $
    \EndFor
    \State $y \gets \text{Reshape}(Y) \in \mathbb{R}^{mn \times 1}$ \Comment{Vectorize the matrix}
    \State \Return $y$
\end{algorithmic}
\label{algo:vgx}
\end{algorithm}

\noindent The sensitivities $\frac{\partial F}{\partial x_k} v$ and $v^T \frac{\partial F}{\partial x_k}$, for some vector $v \in \mathbb{R}^n$, can be calculated efficiently using a single pass of forward mode and reverse mode automatic differentiation across $F$ respectively. To calculate the sensitivities across $F$, we use the discretize-then-optimize approach, however, the optimize-then-discretize approach can also be used.  

\begin{algorithm}
\caption{Computing gradients}
\begin{algorithmic}
    \Require vector $v \in \mathbb{R}^{p \times 1}$
    \Function{Hvp}{$v$}
    \State \Return \Call{Algorithm \ref{algo:gxv}}{}(\Call{Algorithm \ref{algo:vgx}}{$v$}) + \Call{Algorithm \ref{algo:gpv}}{} (\Call{Algorithm \ref{algo:vgp}}{$v$})
    \EndFunction
    \State $\Delta \lambda \gets $ LinearSolve(Hvp, $G$ - \Call{Algorithm \ref{algo:vgx}}{$L_x$} - \Call{Algorithm \ref{algo:vgp}}{$L_p$} ) \Comment{LinSolve using CG or Newton}
    \State $\Delta p \gets - L_x - $ \Call{Algorithm \ref{algo:vgx}}{$\Delta \lambda$} \Comment{Equation \ref{eqn:p_update}}
    \State $\Delta x \gets - L_p - $ \Call{Algorithm \ref{algo:vgp}}{$\Delta \lambda$} \Comment{Equation \ref{eqn:x_update}}
    \State \Return $\Delta p, \Delta x, \Delta \lambda$
\end{algorithmic}
\label{algo:backward}
\end{algorithm}

\subsection{Exploiting sparsity of Jacobian}

\begin{figure}[h!]
    \centering
    % [trim={left bottom right top},clip]
    \includegraphics[width = \linewidth, height = 0.28\textheight, trim = 40 180 40 130, clip]{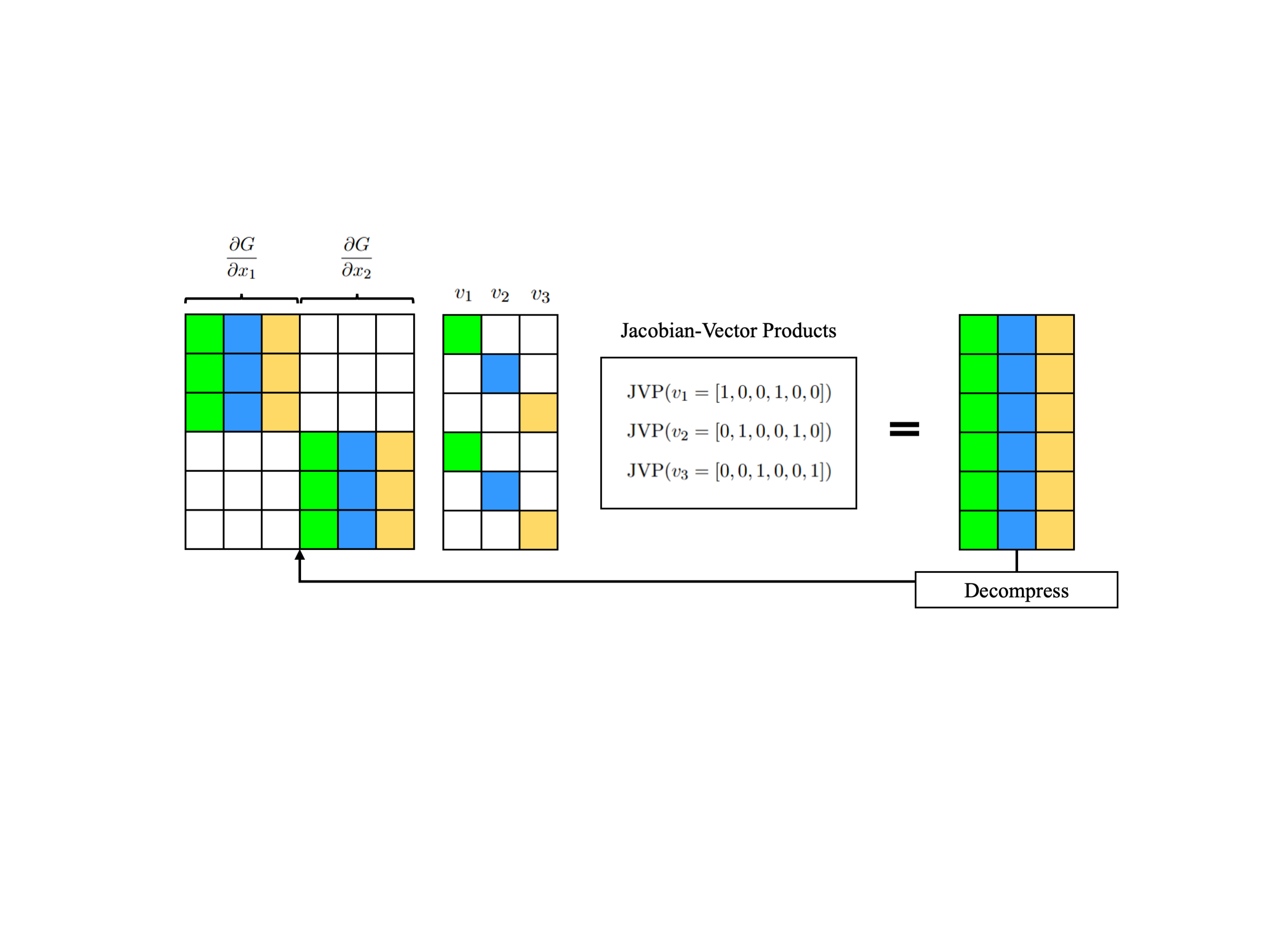}
    \caption{The Jacobian $G_x$ can be computed in only $m$ JVP calls by exploiting its sparsity structure, compared to $mn$ JVP calls required by naive approach}
    \label{fig:jvp}
\end{figure}

For smaller problems, we observe that computing the Jacobians ($G_p, \ G_x$) and solving the linear system is much faster than using the conjugate gradient method for equations \ref{eqn:lambda_update} and \ref{eqn:final_update}. In such cases, we want to compute the Jacobians faster by exploiting their sparsity pattern. Since the Jacobian $G_x$, defined in equation \ref{eqn:jac_x}, is a square matrix, we use the Jacobian-vector product (JVP or forward-mode automatic differentiation) to compute the Jacobian, since it is more memory efficient than the vector-Jacobian product (VJP or reverse-mode automatic differentiation) \cite{griewank2008evaluating}. Naively, the number of JVP's required to construct the Jacobian is proportional to both $m$ and $n$. However, by exploiting the sparsity structure, as shown in the figure \ref{fig:jvp} for the case $m = 2, \ n = 3$, the number of JVP's required is only proportional to $n$. 

\begin{equation}\label{eqn:jac_x}
    G_x = 
    \begin{bmatrix}
        I &  &  & \\[6pt]
        -\frac{\partial F_1}{\partial x_1} & I & & \\[6pt]
         & \ddots & \ddots & \\[6pt]
         & & - \frac{\partial F_{m - 1}}{\partial x_{m - 1}} & I \\[6pt]
    \end{bmatrix} _{mn \times mn} = \quad I_{mn \times mn} \ + \ \underbrace{\begin{bmatrix}
         &  &  & \\[6pt]
        - \frac{\partial F_1}{\partial x_1} & & & \\[6pt]
         & \ddots & & \\[6pt]
         & & - \frac{\partial F_{m - 1}}{\partial x_{m - 1}} & \\[6pt]
    \end{bmatrix} }_{\text{exploiting sparsity}}
\end{equation}

Unfortunately, the Jacobian $G_p$, defined in equation \ref{eqn:jac_p}, is dense, so its sparsity structure cannot be exploited. However, in the case of neural networks, where $p \gg mn$, we compute the Jacobian using $mn$ VJP's instead of $p$ JVP's \citep{griewank2008evaluating}.

\begin{equation}\label{eqn:jac_p}
    G_p = \begin{bmatrix}
          0 \\[6pt]
        - \frac{\partial F_1}{\partial p} \\[6pt]
         \vdots \\[6pt]
        - \frac{\partial F_{m - 1}}{\partial p} \\[6pt]
    \end{bmatrix}_{mn \times p}
\end{equation}

\section{Experiments}

\begin{figure}[h!]
    \centering
    % [trim={left bottom right top},clip]
    \includegraphics[width = \linewidth, height = 0.4\textheight, trim = 20 90 40 100, clip]{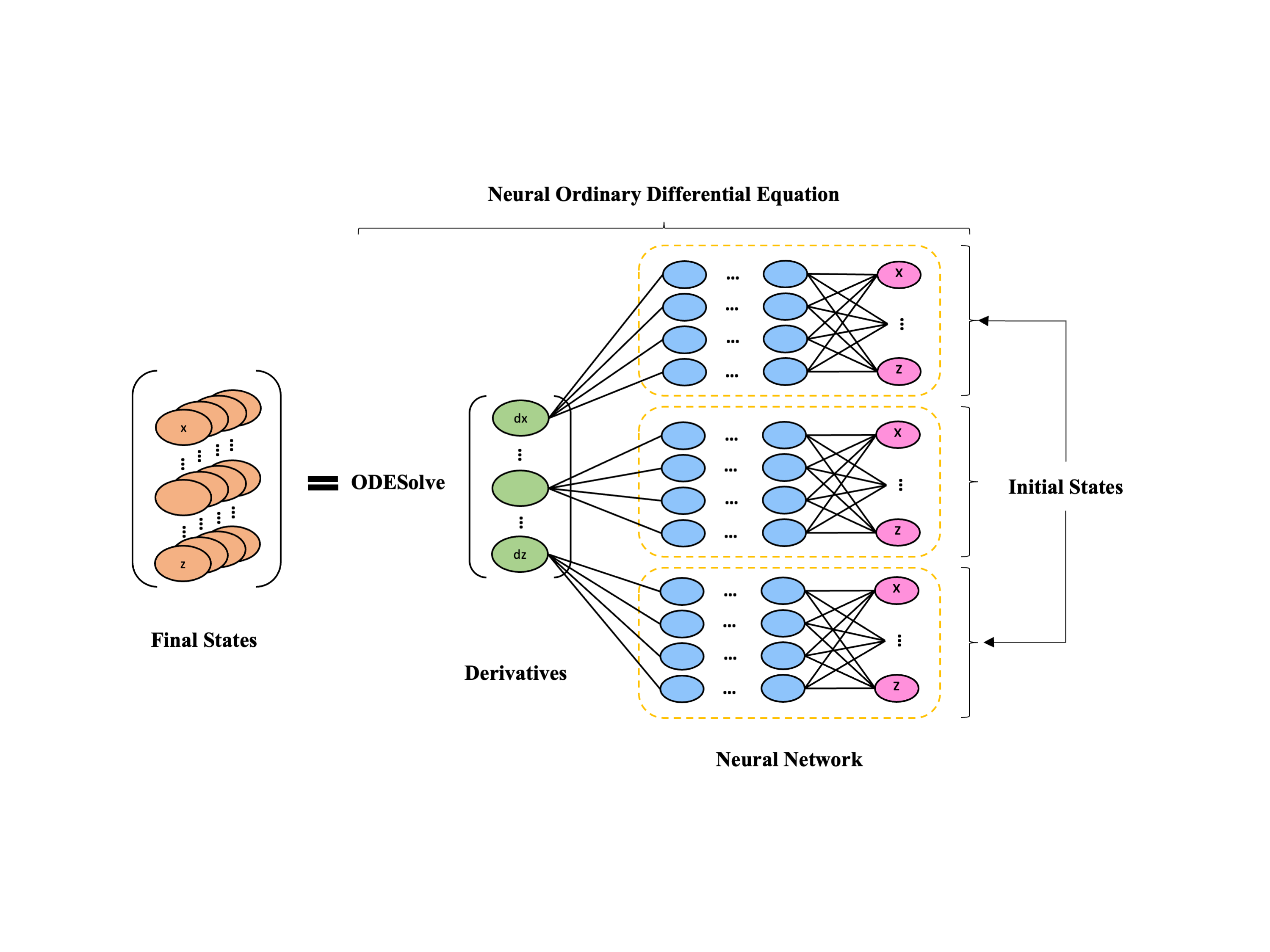}
    \caption{A pictorial representation of the neural network architecture used for training multiple-shooting neural ordinary differential equations}
    \label{fig:nn}
\end{figure}

In this section, we train a MS-NODE using data generated from several oscillatory dynamical systems. A pictorial representation of the neural network architecture is shown in Figure \ref{fig:nn}. The ODE solver requires derivatives with respect to time for each state, which are provided by separate neural networks—one for each state variable. The input dimension of each neural network corresponds to the dimension of the state vector, while the output is scalar. The dimensions of the hidden layers for each system are summarized in Table \ref{tab:training_results}. 

The initial guess of the states at each of the shooting intervals is assumed as the initial measurement. After computing the gradients corresponding to the states at each of the intervals $\Delta x$, parameters $\Delta p$ and the Lagrange variables $\Delta \lambda$ using Algorithm \ref{algo:backward}, we use the Adam optimizer \citep{Kingma2014AdamAM} to get the corresponding updates. The code is availabe at \url{https://github.com/siddharth-prabhu/MS-NODE}. We observe that, for most systems, starting with a relatively high learning rate of $0.01$ and gradually decreasing it as the solution converges helps prevent the model from converging to trivial solutions. Once the MS-NODE model has been trained, we simulate the model in a single-shooting approach on unseen test data. The performance of the models is evaluated using mean squared error (MSE), which is summarized in Table \ref{tab:training_results}, for different systems, for both training and testing data. Additionally, we report the number of shooting intervals chosen, the number of epochs required, and the infinity norm of constraint violation at convergence (stopping point).

\begin{table}[ht]
\centering
\begin{tabular}{lcccccccc}
\hline
\multirow{3}{*}{System} & \multirow{3}{*}{\centering $n$} & \multirow{3}{4em}{\centering Hidden Layers} & \multirow{3}{*}{\centering Intervals} & \multirow{3}{3em}{\centering Learning Rate} & \multirow{3}{*}{Epochs} & \multirow{3}{5em}{\centering Train Loss $(\times 10 ^{-4})$} & \multirow{3}{4em}{\centering $|G|_{\infty}$ $(\times 10 ^{-4})$} & \multirow{3}{4em}{\centering Test Loss $(\times 10 ^{-4})$}\\
& & & & &  & \\
& & & & &  & \\
\hline 
Lotka Volterra & 2 & [32, 64, 32] & 20 & 0.01 & 400 & 0.1 & 1.4 & 0.1 \\
Van der Pol & 2 & [32, 64, 64] & 20 & $0.01^*$ & 2500 & 0.19 & 5.78 & 0.34 \\
FH-Nagumo$^{**}$ & 2 & [32] & 20 & 0.01 & 700 & 0.99 & 1.16 & 1.18 \\
Goodwin & 3 & [32, 64, 32] & 20 & 0.01 & 420 & 0.05 & 2.17 & 0.31 \\
Brusselator & 2 & [32, 64, 64, 128] & 40 & 0.01 & 850 & 0.42 & 3.63 & 1.2 \\
Zebrafish$^{**}$ & 2 & [32, 64, 32, 16] & 100 & $0.01^*$ & 1900 & 0.45 & 5.17 & 11.8 \\
Oregonator$^{**}$ & 3 & [32, 64, 64, 128] & 20 & $0.01^*$ & 5000 & 4.26 & 22.03 & 5.57 \\
MHD$^{**}$ & 6 & [32, 64, 64] & 20 & $0.01^*$ & 1700 & 2.63 & 1.84 & - \\
KM$^{**}$ & 3 & [32, 64, 64] & 20 & $0.01^*$ & 1500 & 0.36 & 5.73 & - \\
Calcium Ion$^{**}$ & 4 & [32, 64, 128, 16] & 20 & $0.01^*$ & 3000 & 1.89 & 4.69 & - \\
\hline
\multicolumn{7}{l}{* We reduce the learning rate as the solution converges} \\
\multicolumn{7}{l}{** We scale the data before training} \\ \\
\end{tabular}
\caption{Summary of training conditions and model performance (mean squared error) on training and testing data generated from different systems}
\label{tab:training_results}
\end{table}

\subsection{Lotka Volterra System}
\begin{figure}[h!]
    \centering
    % [trim={left bottom right top},clip]
    \includegraphics[width = \linewidth, height = 0.25\textheight, trim = 20 150 20 150, clip]{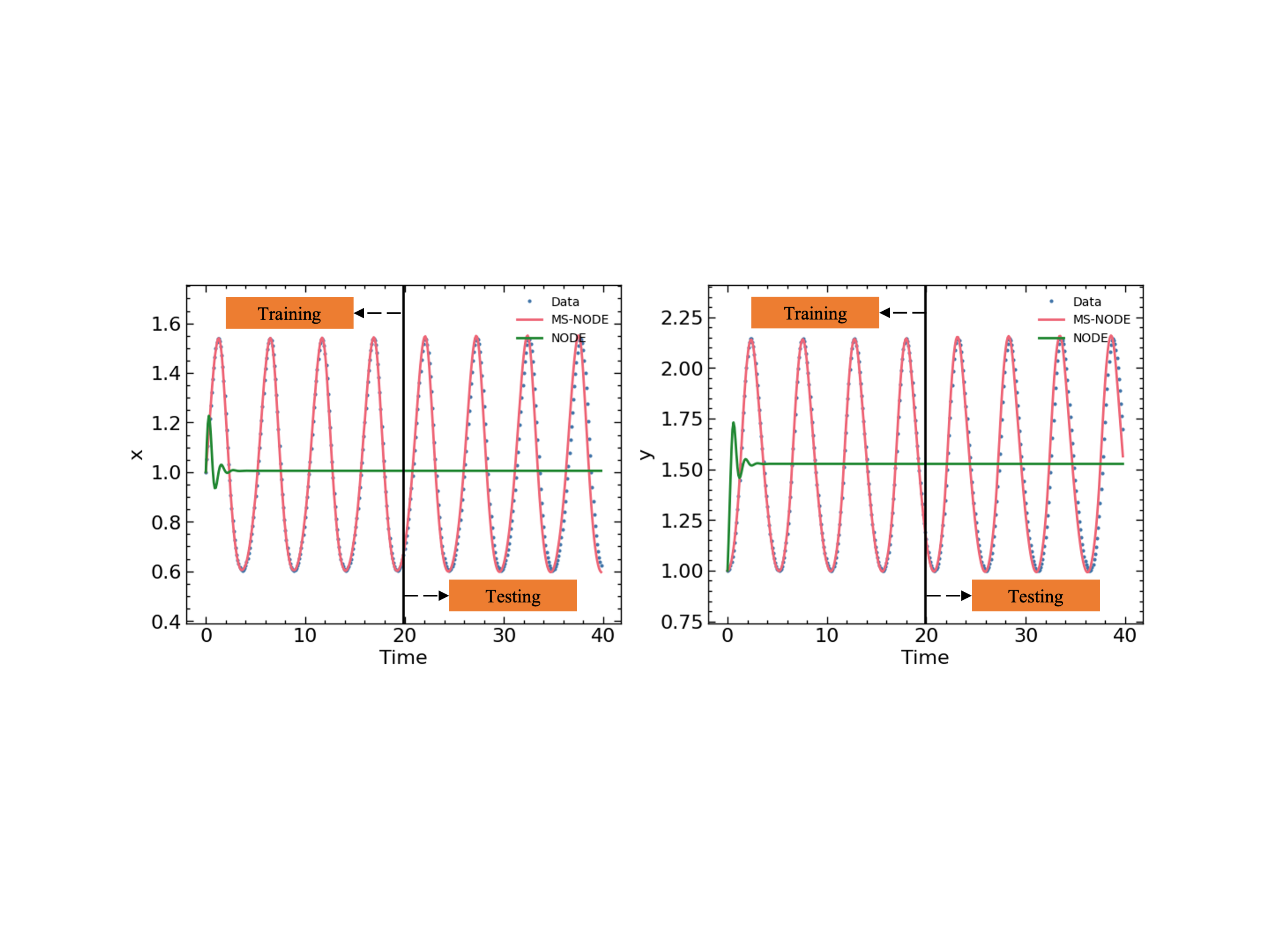}
    \caption{Comparison between the model trained using multiple-shooting (red) and the model trained using single-shooting (green) on both training and testing data (blue) generated from the Lotka Volterra system.}
    \label{fig:LV}
\end{figure}

This system of ODEs, given in equation \ref{eqn:LV}, models the interaction of a predator and its prey \citep{wangersky1978lotka}. The initial conditions are chosen to be $x(t = 0) = 1, \ y(t = 0) = 1$. The model is simulated from $t_i = 0$ to $t_f = 20$ (sec), and measurements are collected every $0.1$ seconds. Figure \ref{fig:LV} compares the performance of the model trained using multiple-shooting to that of the model trained using single-shooting. 

\begin{equation}\label{eqn:LV}
\begin{aligned}
    \frac{dx}{dt} & = 1.5 x - x y \\
    \frac{dy}{dt} & = - y + x y
\end{aligned}
\end{equation}

\subsection{Goodwin System}

\begin{figure}[h!]
    \centering
    % [trim={left bottom right top},clip]
    \includegraphics[width = \linewidth, height = 0.25\textheight, trim = 20 170 20 170, clip]{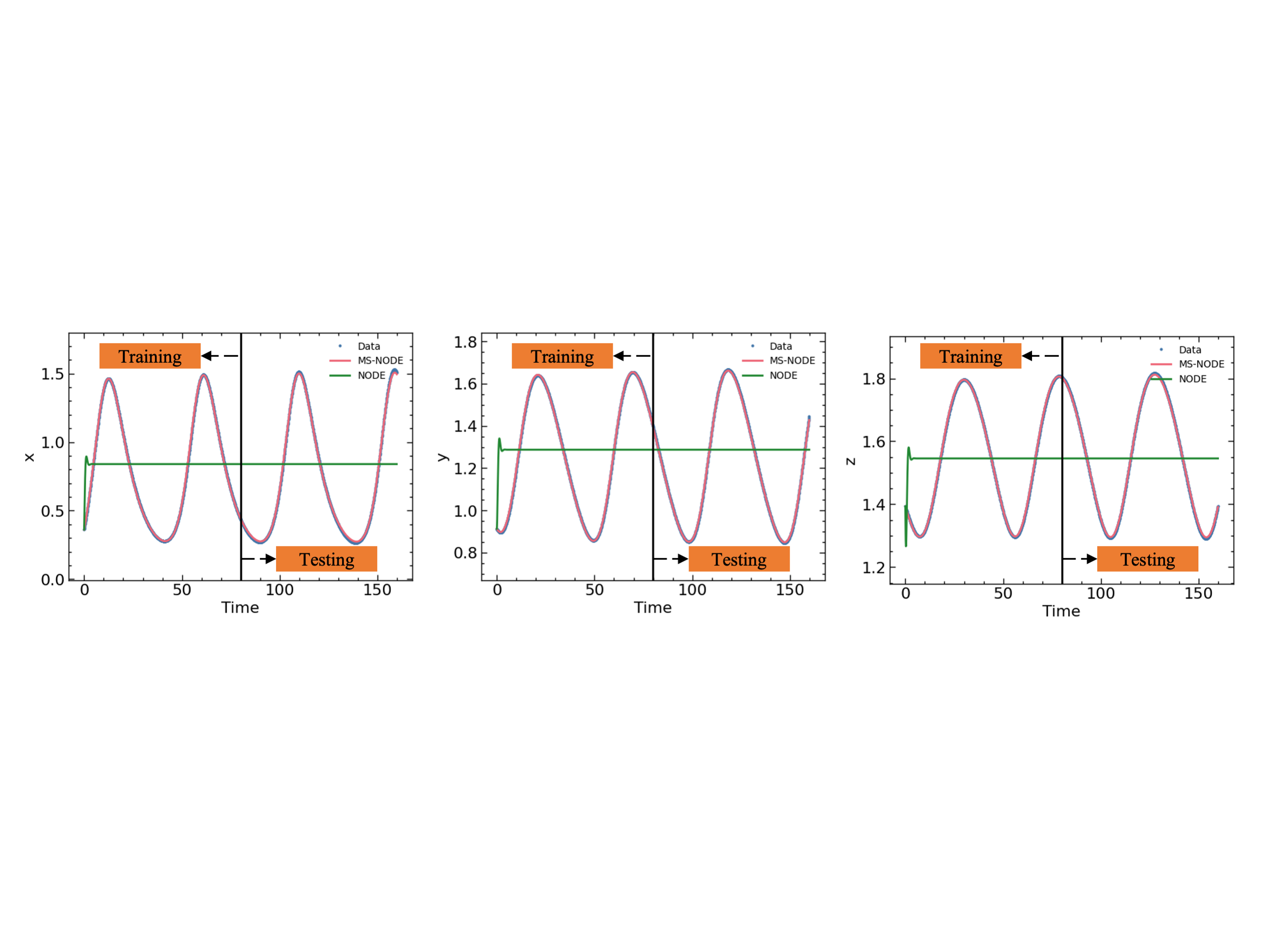}
    \caption{Comparison between the model trained using multiple-shooting (red) and the model trained using single-shooting (green) on both training and testing data (blue) generated from the Goodwin system.}
    \label{fig:GW}
\end{figure}

This system of ODEs, given in equation \ref{eqn:GW}, models a biological oscillator, which has been applied to enzyme kinetics \citep{goodwin1965oscillatory}. The parameters are set to $a = 3.4884, \ A = 2.15, \ b = 0.0969, \ \alpha = 0.0969, \ \beta = 0.0581, \ \gamma = 0.0969, \ \sigma = 10, \ \delta = 0.0775 $. The initial conditions are chosen to be $x(t = 0) = 0.3617, \ y(t = 0) = 0.9137, \ z(t = 0) = 1.3934$. The model is simulated from $t_i = 0$ to $t_f = 80$(sec), and measurements are collected every $0.1$ seconds. Figure \ref{fig:GW} compares the performance of the model trained using multiple-shooting to that of the model trained using single-shooting. 

\begin{equation}\label{eqn:GW}
\begin{aligned}
    \frac{dx}{dt} & = \frac{a}{A + z^{\sigma}} - b x \\
    \frac{dy}{dt} & = \alpha x - \beta y \\
    \frac{dz}{dt} & = \gamma y - \delta z
\end{aligned}
\end{equation}

\subsection{Van der Pol System}

\begin{figure}[h!]
    \centering
    % [trim={left bottom right top},clip]
    \includegraphics[width = \linewidth, height = 0.25\textheight, trim = 20 150 20 150, clip]{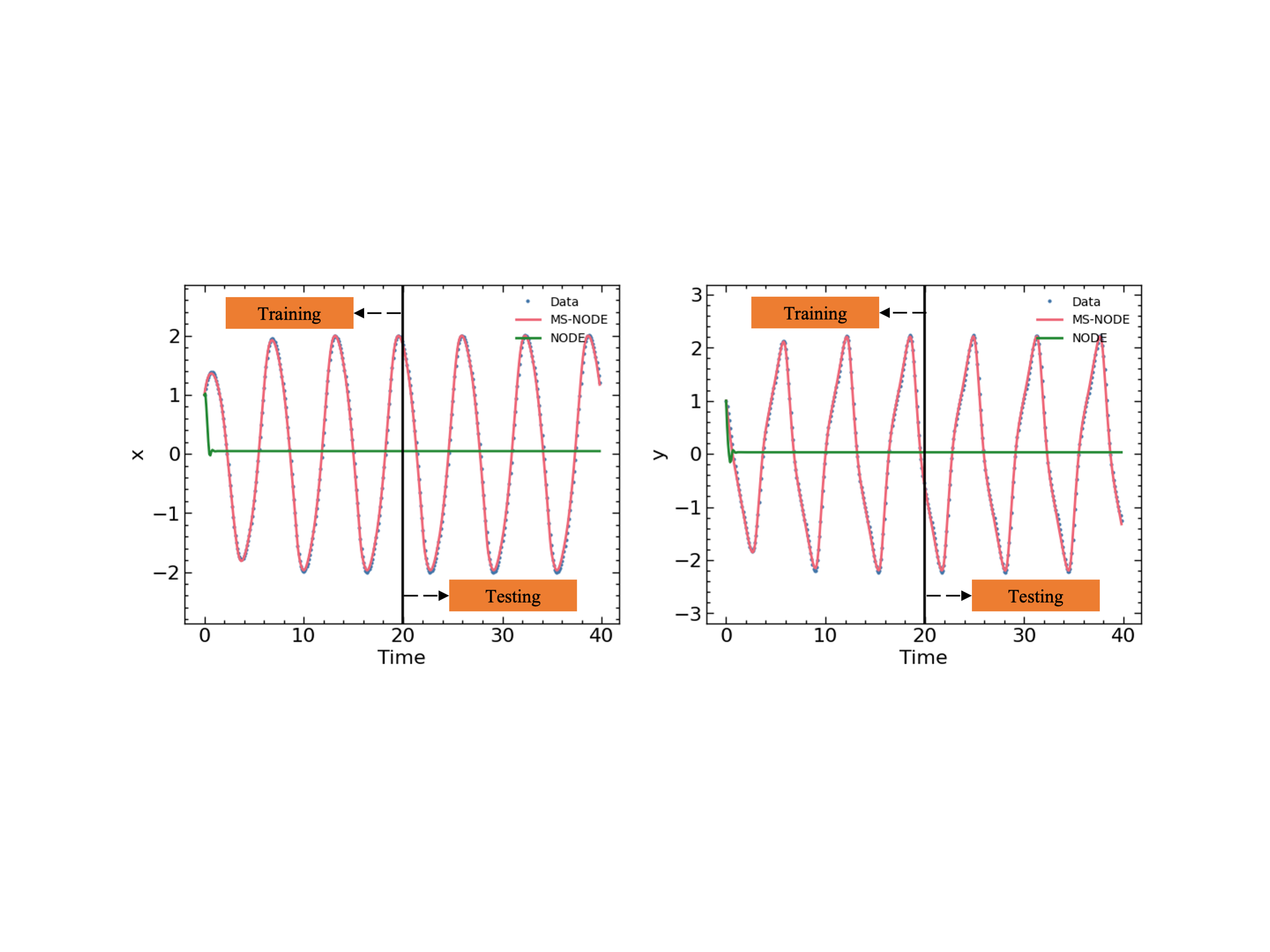}
    \caption{Comparison between the model trained using multiple-shooting (red) and the model trained using single-shooting (green) on both training and testing data (blue) generated from the Van der Pol system.}
    \label{fig:VDP}
\end{figure}

Equation \ref{eqn:VDP} describes the behavior of non-conservative, oscillating system with nonlinear damping \citep{guckenheimer2003dynamics}. The initial conditions are given as $x(t = 0) = 1$ and $y(t = 0) = 1$. The model is simulated from $t_i = 0$ to $t_f = 20$(sec), and measurements are collected every $0.1$ seconds. Figure \ref{fig:VDP} compares the performance of the model trained using multiple-shooting to that of the model trained using single-shooting. 

\begin{equation}\label{eqn:VDP}
\begin{aligned}
    \frac{dx}{dt} & = x \\
    \frac{dy}{dt} & = 0.5 (1 - x^2) y - x
\end{aligned}
\end{equation}

\subsection{FitzHugh-Nagumo System}

\begin{figure}[h!]
    \centering
    % [trim={left bottom right top},clip]
    \includegraphics[width = \linewidth, height = 0.25\textheight, trim = 20 150 20 150, clip]{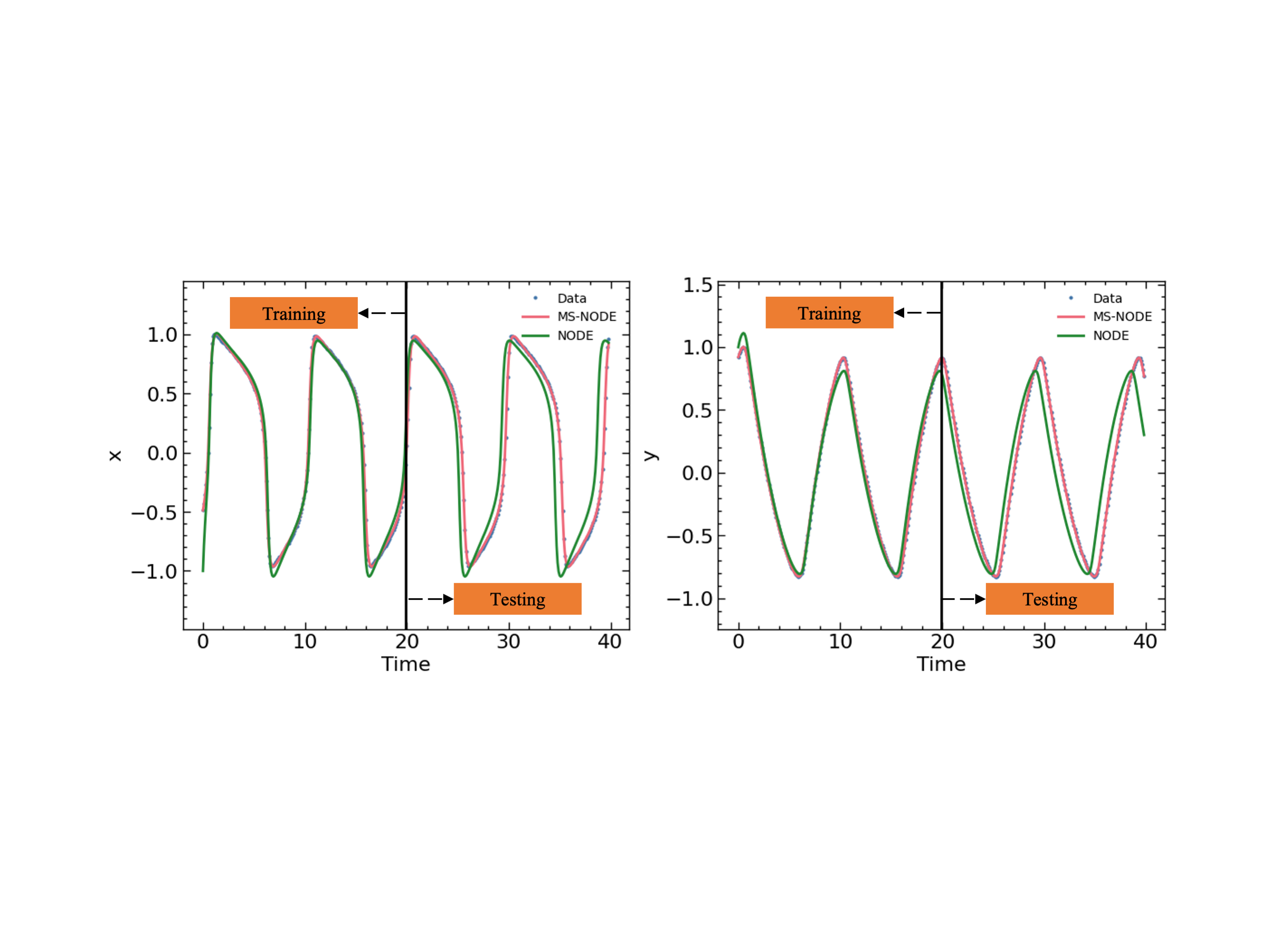}
    \caption{Comparison between the model trained using multiple-shooting (red) and the model trained using single-shooting (green) on both training and testing data (blue) generated from the FitzHugh-Nagumo system.}
    \label{fig:FHN}
\end{figure}

Equation \ref{eqn:FHN} models the potential of squid neurons \citep{fitzhugh1961impulses, calver2019parameter}. The initial conditions and parameters are set to $x(t = 0) = -1, \ y(t = 0) = 1, \ a = 0.2, \ b = 0.2, \ $ and $c = 3.5$. The model is simulated from $t_i = 0$ to $t_f = 20$(sec), and measurements are collected every $0.1$ seconds. Figure \ref{fig:FHN} compares the performance of the model trained using multiple-shooting to that of the model trained using single-shooting. 

\begin{equation}\label{eqn:FHN}
\begin{aligned}
    \frac{dx}{dt} & = c\left( x - \frac{x^3}{3} + y \right) \\
    \frac{dy}{dt} & = \frac{- (x - a + b y) }{c}
\end{aligned}
\end{equation}

\subsection{Brusselator Chemical Reaction System}

\begin{figure}[h!]
    \centering
    % [trim={left bottom right top},clip]
    \includegraphics[width = \linewidth, height = 0.25\textheight, trim = 20 150 20 150, clip]{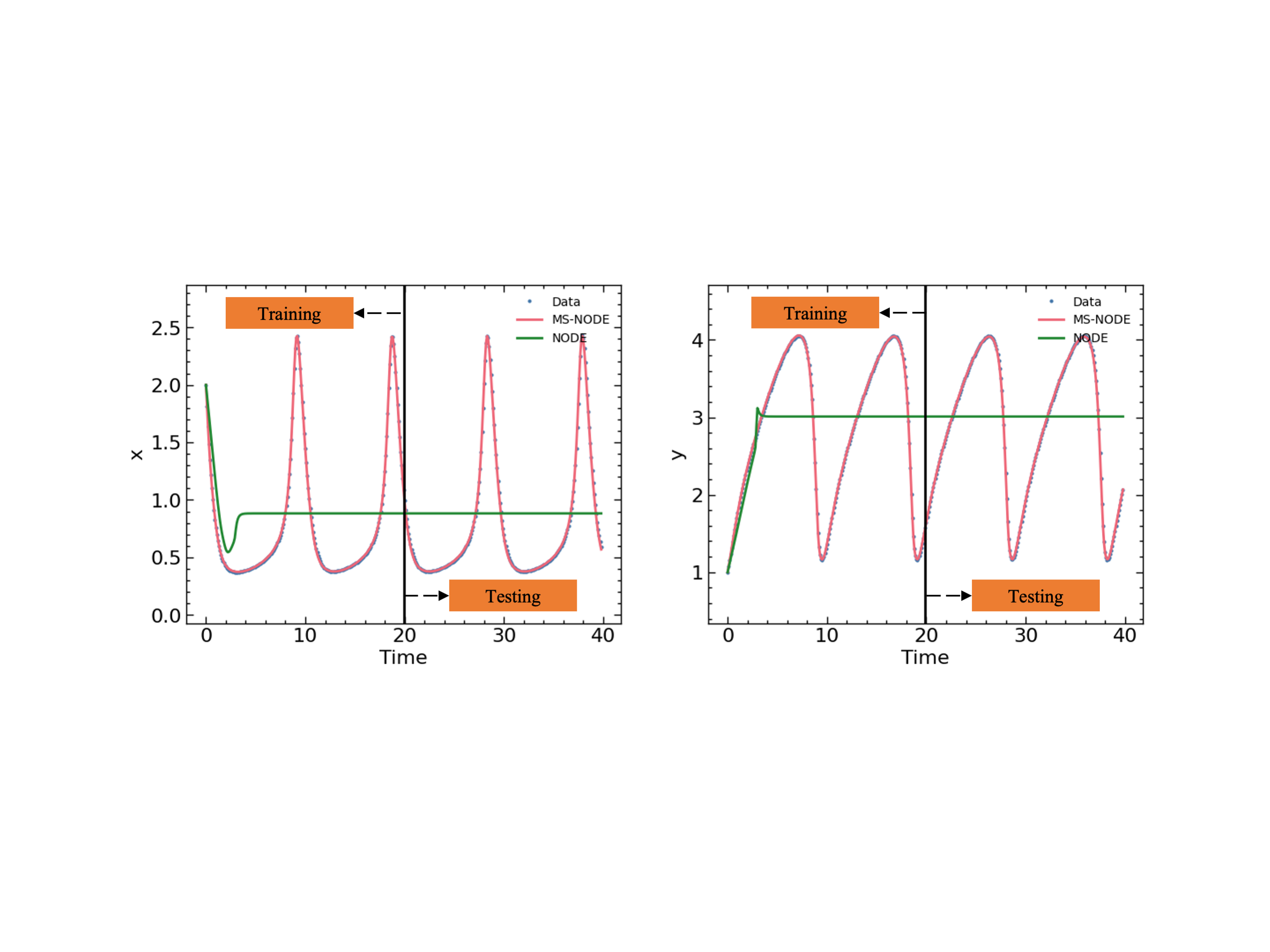}
    \caption{Comparison between the model trained using multiple-shooting (red) and the model trained using single-shooting (green) on both training and testing data (blue) generated from the Brusselator Chemical Reaction system.}
    \label{fig:KOSC}
\end{figure}

Equation \ref{eqn:KOSC} forms a system of autocatalytic oscillating reactions proposed in \cite{lefever1971chemical}. The parameters are set to $a = 0.8, \ b = 2, \ c = 0.8 $. The initial conditions are given as $x(t = 0) = 2$ and $y(t = 0) = 1$. The model is simulated from $t_i = 0$ to $t_f = 20$(sec), and measurements are collected every $0.1$ seconds. Figure \ref{fig:KOSC} compares the performance of the model trained using multiple-shooting to that of the model trained using single-shooting. 

\begin{equation}\label{eqn:KOSC}
\begin{aligned}
    \frac{dx}{dt} & = a - (b + 1) x + c x^2 y \\
    \frac{dy}{dt} & = b x - c x^2 y
\end{aligned}
\end{equation}

\subsection{Zebrafish System}

\begin{figure}[h!]
    \centering
    % [trim={left bottom right top},clip]
    \includegraphics[width = \linewidth, height = 0.25\textheight, trim = 20 150 20 150, clip]{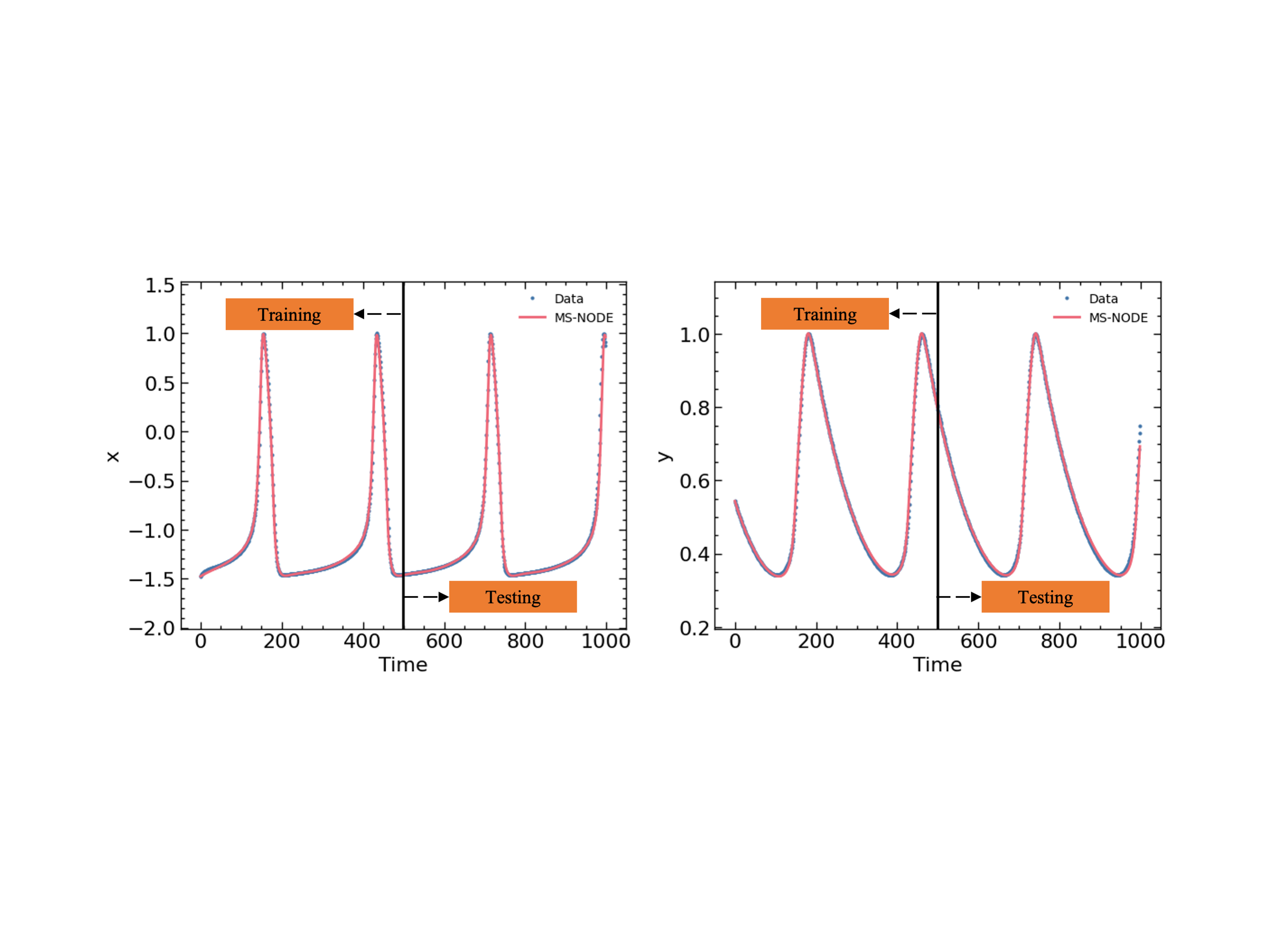}
    \caption{Comparison between the model trained using multiple-shooting (red) and the model trained using single-shooting (green) on both training and testing data (blue) generated from the Zebrafish system.}
    \label{fig:ZF}
\end{figure}

This model \citep{dattner2015model, calver2019parameter} describes the behavior of a neuron and is given by equation \ref{eqn:ZF}. The parameters are set to $a_1 = 0.7934, \ a_2 = 0.0411, \ p_1 = 5, \ p_2 = 2.86 \times 10^{-1}, \ p_3 = -5.095 \times 10^{-3}, \ p_4 = -3.748 \times 10^{-4}, \ p_5 = -1.255 \times 10^{-1}, \ p_6 = -5.919 \times 10^{-3}, \ p_7 = -5.737 \times 10^{-3}$. The initial conditions are set to $x(t = 0) = -20.5693$ and $y(t = 0) = 28.1786$. The model is simulated from $t_i = 0$ to $t_f = 500$(sec), and measurements are collected every $1$ second. Figure \ref{fig:ZF} compares the performance of the model trained using multiple-shooting to that of the model trained using single-shooting. 

\begin{equation}\label{eqn:ZF}
\begin{aligned}
        \frac{dx}{dt} & = p_1 + p_2 x + p_3 x^2 + p_4 x^3 + p_5 y + p_6 xy \\
        \frac{dy}{dt} & = a_1 + a_2 x + p_7 y
\end{aligned}
\end{equation}

\subsection{Oregonator System}
The Oregonator, Equation \ref{eqn:ORG}, is the simplest realistic model of the autocatalytic Belousov-Zhabotinsky chemical reaction \cite{gray2002analysis}.  

\begin{equation}\label{eqn:ORG}
\begin{aligned}
        \frac{dx}{dt} & = \frac{1}{\epsilon} \left( x(1 - x) - fy \frac{x - q}{z + q} \right) \\
        \frac{dy}{dt} & =  x - y\\
        \frac{dz}{dt} & = \phi (y - z)
\end{aligned}
\end{equation}

\begin{figure}[h!]
    \centering
    % [trim={left bottom right top},clip]
    \includegraphics[width = \linewidth, height = 0.25\textheight, trim = 0 150 0 150, clip]{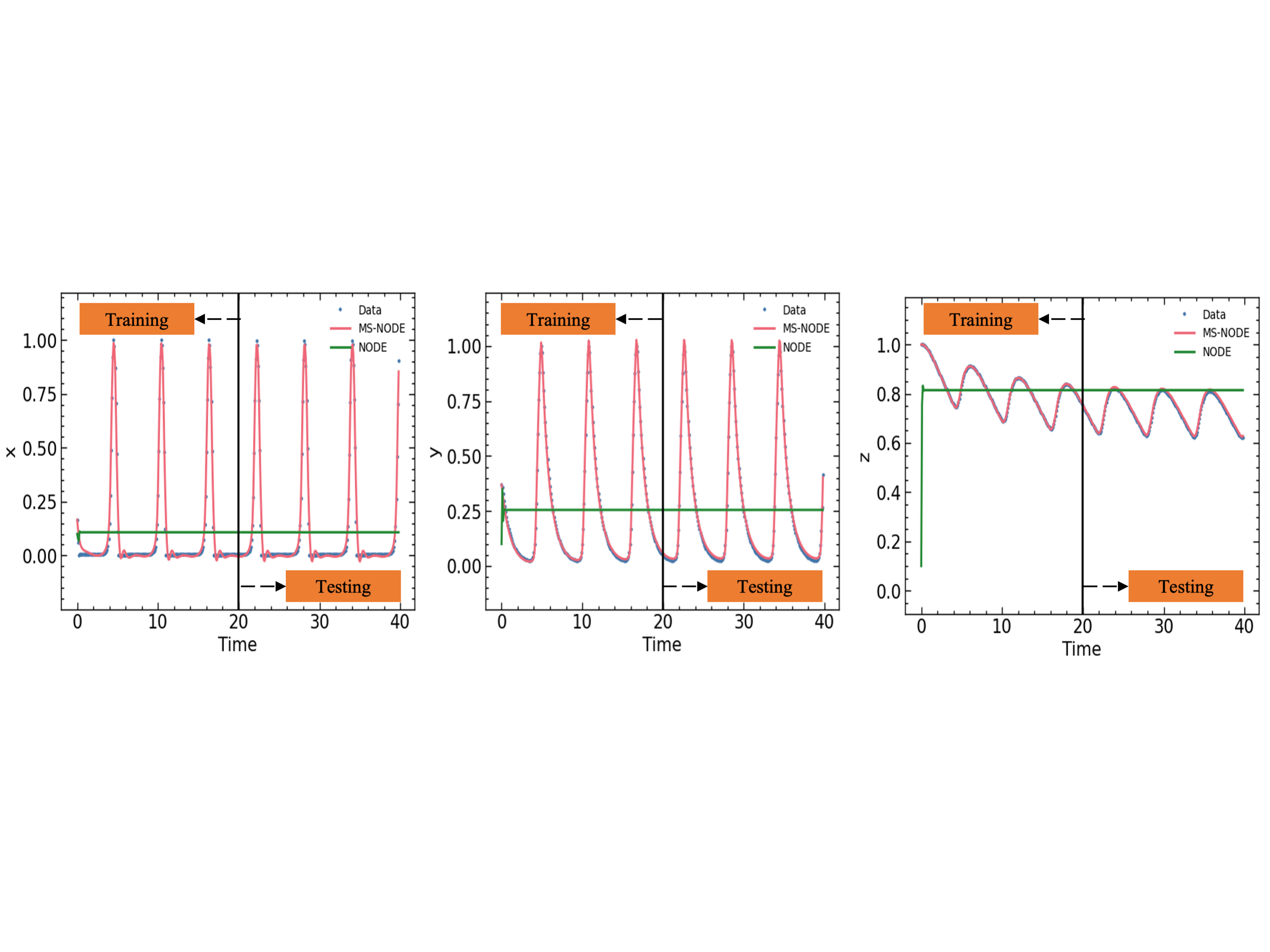}
    \caption{Comparison between the model trained using multiple-shooting (red) and the model trained using single-shooting (green) on both training and testing data (blue) generated from the Oregonator system.}
    \label{fig:ORG}
\end{figure}

The parameters are set to $\epsilon = 0.1, \ f = 1.4, \ q = 0.002, \ \phi = 0.1$. With initial conditions as $x(t = 0) = 0.1, \ y(t = 0) = 0.1, \ z(t = 0) = 0.1$, the equations are simulated from $t_i = 0$ to $t_f = 20$(sec), and measurements are collected every $0.1$ seconds. Figure \ref{fig:ORG} compares the performance of the model trained using multiple-shooting to that of the model trained using single-shooting. 

\subsection{MHD System}

Magnetohydrodynamic \cite{kaptanoglu2021promoting} is a highly oscillatory system with dynamics given in equation \ref{eqn:MHD}. 

\begin{equation}\label{eqn:MHD}
\begin{aligned}
        \frac{d}{dt}\begin{bmatrix}
            x \\
            y \\
            z \\
            w \\
            a \\
            b
        \end{bmatrix} = \begin{bmatrix}
            -2 v & 0 & 0 & 0 & 0 & 0 \\
            0 & -5 v & 0 & 0 & 0 & 0 \\
            0 & 0 & -9 v & 0 & 0 & 0 \\
            0 & 0 & 0 & -2 \mu & 0 & 0 \\
            0 & 0 & 0 & 0 & -5 \mu & 0 \\
            0 & 0 & 0 & 0 & 0 & -9 \mu \\
        \end{bmatrix} \begin{bmatrix}
            x \\
            y \\
            z \\
            w \\
            a \\
            b
        \end{bmatrix} + \begin{bmatrix}
            4 (yz - ab) \\
            -7 (xz - wb) \\
            3(xy - wa) \\
            2(by - za) \\
            5(zw - bx) \\
            9(xa - wy)
        \end{bmatrix}
\end{aligned}
\end{equation}

\begin{figure}[h!]
    \centering
    % [trim={left bottom right top},clip]
    \includegraphics[width = \linewidth, height = 0.42\textheight, trim = 0 70 0 70, clip]{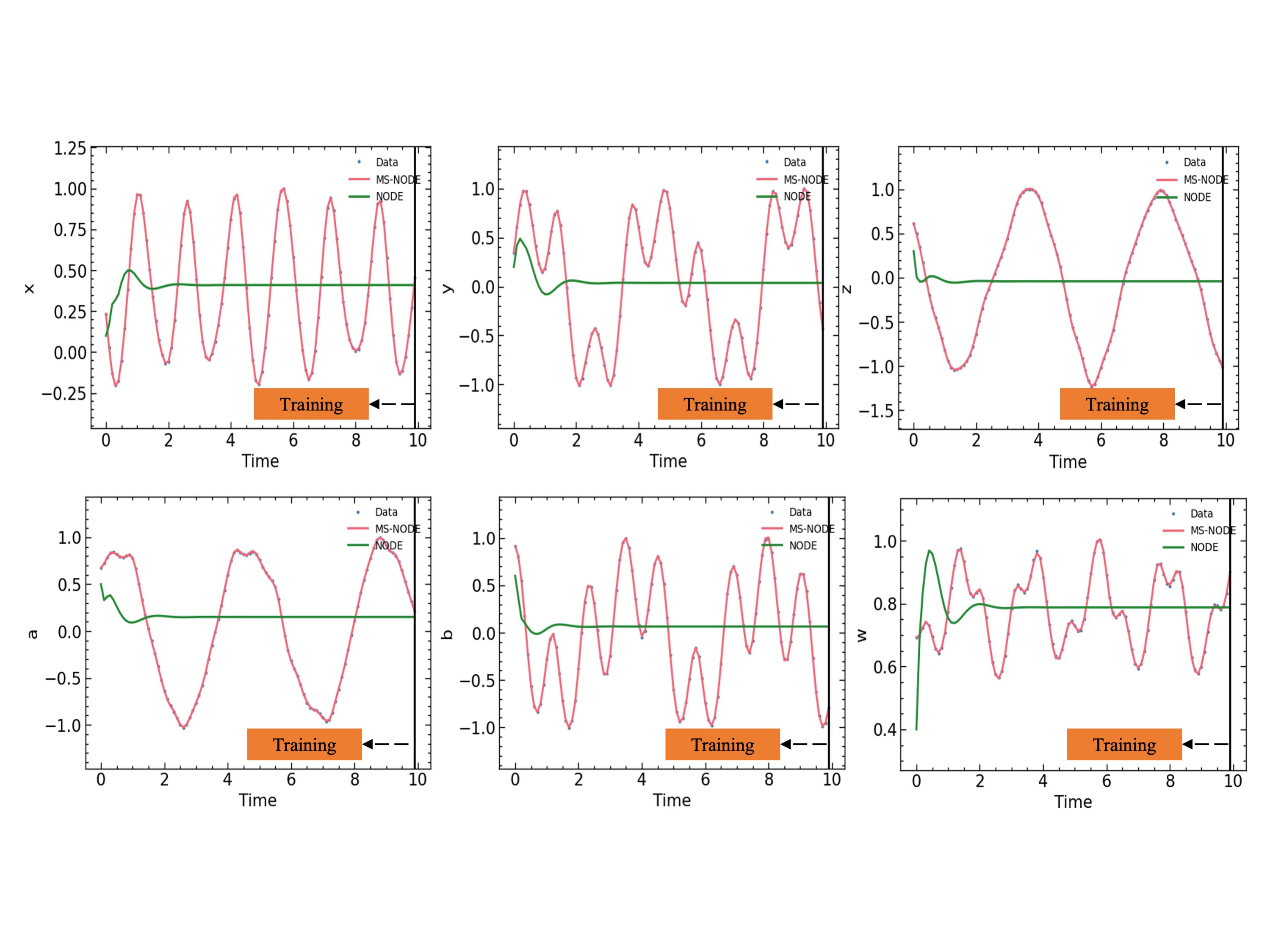}
    \caption{Comparison between the model trained using multiple-shooting (red) and the model trained using single-shooting (green) on training data (blue) generated from the MHD system.}
    \label{fig:MHD}
\end{figure}

The parameters are set to $v = 0, \ \mu = 0$. With initial conditions as $x(t = 0) = 0.1, \ y(t = 0) = 0.2, \ z(t = 0) = 0.3, \ w(t = 0) = 0.4, \ a(t = 0) = 0.5, \ b(t = 0) = 0.6$, the equations are simulated from $t_i = 0$ to $t_f = 10$(sec), and measurements are collected every $0.1$ seconds. Figure \ref{fig:MHD} shows the performance of the model trained using multiple-shooting to that of the model trained using single-shooting. We observe that the model fits the training data but does not generalize to the testing data. 

\subsection{KM System}
The Kermack-McKendrick model given by equation \ref{eqn:KM}, is a system of delayed differential equations that models the spread of a disease within a population \cite{calver2019parameter}. 

\begin{equation}\label{eqn:KM}
\begin{aligned}
        \frac{dx}{dt} & = - x(t) y(t - \tau _1) + y(t - \tau _2) \\
        \frac{dy}{dt} & = x(t) y(t - \tau _1) - y(t) \\
        \frac{dz}{dt} & = y(t) - y(t - \tau _2)
\end{aligned}
\end{equation}

\begin{figure}[h!]
    \centering
    % [trim={left bottom right top},clip]
    \includegraphics[width = \linewidth, height = 0.25\textheight, trim = 0 160 0 160, clip]{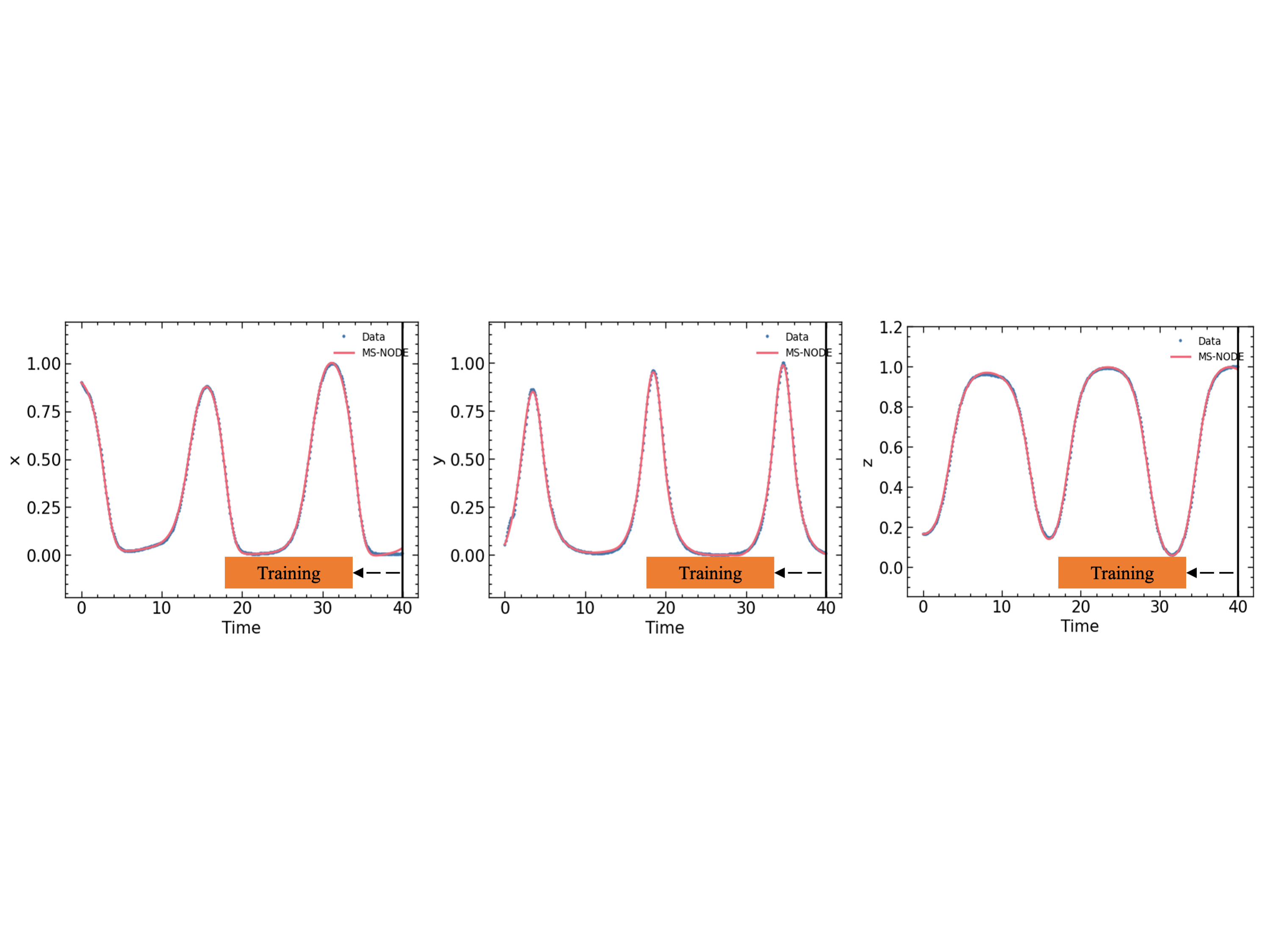}
    \caption{Performance of the model trained using multiple-shooting (red) on training (blue) generated from the KM system.}
    \label{fig:KM}
\end{figure}

The parameters are set to $\tau _1 = 1, \ \tau _2 = 10$. With initial conditions as $x(t \leq 0) = 5, \ y(t \leq 0) = 0.1, \ z(t \leq 0) = 1$, the equations are simulated from $t_i = 0$ to $t_f = 40$(sec), and measurements are collected every $0.1$ seconds. To account for the delayed terms, time is concatenated with the states and passed as input to the neural network. With this modification, although the model captures the training data correctly, the neural networks ability to generalize to the testing data is lost. Figure \ref{fig:KM} shows the performance of the model trained using multiple-shooting. In this figure, we do not show the performance of the model trained using single-shooting because the trajectory explodes.  

\subsection{Calcium Ion System}
We use the oscillatory dynamics of calcium ion in the eukaryotic cells described in \cite{kummer2000switching, calver2019parameter}. The dynamics consists of four differential equations given in equation \ref{eqn:CAL}

\begin{equation}\label{eqn:CAL}
\begin{aligned}
    \frac{dx}{dt} & = k_1 + k_2 x - k_3 y \frac{x}{x + Km_1} - k_4 z \frac{x}{x + Km_2} \\
    \frac{dy}{dt} & = k_5 x - k_6 \frac{y}{y + Km_3} \\
    \frac{dz}{dt} & = k_7 y z \frac{w}{w + Km_4} + k_8 y + k_9 x - k_{10} \frac{z}{z + Km_5} - k_{11} \frac{z}{z + Km_6} \\
    \frac{dw}{dt} & = - k_7 y z \frac{w}{w + Km_4} + k_{11} \frac{z}{z + Km_6}
\end{aligned}
\end{equation}

The parameters are set to $ k_1 = 0.09, \ k_2 = 2, \ k_3 = 1.27, \ k_4 = 3.73, \ k_5 = 1.27, \ k_6 = 32.24,\  k_7 = 2, \ k_8 = 0.05, \ k_9 = 13.58, \ k_{10} = 153, \ k_{11} = 4.85, \ Km_1 = 0.19, \ Km_2 = 0.73, \ Km_3 = 29.09, \ Km_4 = 2.67, \ Km_5 = 0.16, \ Km_6 = 0.05$. The initial conditions are chosen to be $x(t = 0) = 0.12, \ y(t = 0) = 0.31, \ z(t = 0) = 0.0058, \ w(t = 0) = 4.3$. The model is simulated from $t_i = 0$ to $t_f = 60$(sec), and measurements are collected every $0.1$ seconds. 

\begin{figure}[h!]
    \centering
    % [trim={left bottom right top},clip]
    \includegraphics[width = \linewidth, height = 0.45\textheight, trim = 10 20 10 30, clip]{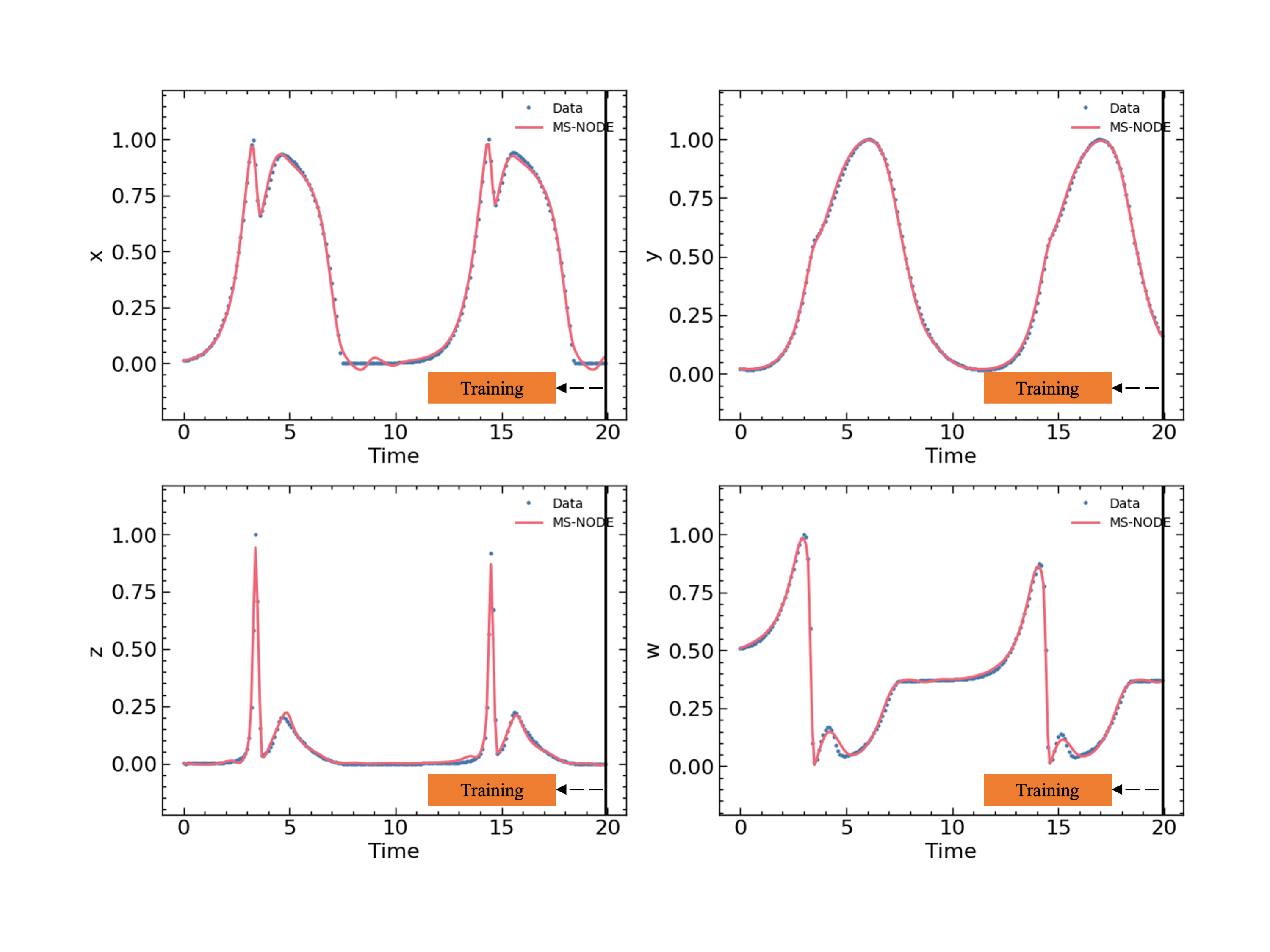}
    \caption{Performance of the model trained using multiple-shooting (red) on training (blue) generated from the Calcium Ion system.}
    \label{fig:CAL}
\end{figure}

\section{Conclusion}
We propose a condensing-based approach to incorporate general shooting equality constraints while training a MS-NODE. We demonstrate the effectiveness of our method on data generated from several oscillatory and complex dynamical equations. Unlike single-shooting training approach, multiple-shooting training approach captures the underlying dynamics (for the given initial conditions) as observed from the model performance on testing data. However, in some cases, such as the KM, MHD, and Calcium Ion systems, MS-NODE accurately captures the training data but fails to generalize to unseen testing data. Despite this, such a model can still be used as a surrogate model and provide a continuous approximation of measurements for parameter estimation \citep{bradley2021two}. This behavior implies that the model has not captured the true underlying physics and is overfitting on training data. In such a case, either stopping early or using high-dimensional neural network may improve generalization.

\clearpage
\bibliographystyle{unsrtnat}
\bibliography{references}
\clearpage

\end{document}